\documentclass[graybox]{svmult}

\usepackage{type1cm}        
\usepackage{makeidx}         
\usepackage{graphicx}        
\usepackage{multicol}        
\usepackage[bottom]{footmisc}

\usepackage{newtxtext}       %
\usepackage{newtxmath}       

\makeindex             

\newcommand{\Err}{\mathcal{L}} 
\usepackage{subcaption}
\newcommand\T{\rule{0pt}{3ex}}       
\newcommand\B{\rule[-2ex]{0pt}{0pt}} 

\usepackage{wrapfig}
\usepackage{diagbox}

\begin{document}

\title*{Introduction to deep learning}
\author{Lihi Shiloh-Perl and Raja Giryes}
\institute{School of Electrical Engineering, Tel Aviv University, e-mail: \{lihishiloh@mail ,raja@tauex\}.tau.ac.il}
%
%
\maketitle

\abstract{Deep Learning (DL) has made a major impact on data science in the last decade. This chapter introduces the basic concepts of this field. It includes both the basic structures used to design deep neural networks and a brief survey of some of its popular use cases.}

\section{General overview}\label{sec:basic_overview}
Neural Networks (NN) have revolutionized the modern day-to-day life. Their significant impact is present even in our most basic actions, such as ordering products on-line via Amazon's Alexa or passing the time with on-line video games against computer agents. 
The NN effect is evident in many more occasions, for example, in medical imaging NNs are utilized for lesion detection and segmentation~\cite{Greenspan2016, ben2016fully}, and tasks such as text-to-speech~\cite{gibiansky2017deep,Sotelo2017Char2WavES} and text-to-image~\cite{Reed_text2image2016} have remarkable improvements thanks to this technology. 
In addition, the advancements they have caused in fields such as natural language processing (NLP)~\cite{devlin2018pretraining,yang2019xlnet_arxiv,liu2020roberta}, optics \cite{Shiloh19Efficient, Haim18Depth}, image processing \cite{Schwartz19DeepISP, Yang19Deep} and computer vision (CV)~\cite{chen2018encoder,gao2018reconet} are astonishing, creating a leap forward in technology such as autonomous driving~\cite{chen2019progressive,Ma2019CVPR}, face recognition~\cite{facenet,CosFace,ArcFace}, anomaly detection~\cite{Kwon2019}, text understanding~\cite{kadlec2016text} and art~\cite{gatys2016image,johnson2016perceptual}, to name a few.  Its influence is powerful and is continuing to grow.

The NN journey began in the mid 1960's with the publication of the Perceptron \cite{Rosenblatt58theperceptron}. Its development was motivated by the formulation of the human neuron activity \cite{McCulloch1943} and research regarding the human visual perception \cite{hubel:single}. 
However, quite quickly, a deceleration in the field was experienced, which lasted for almost three decades. This was mainly the result of lack of theory with respect to the possibility of training the (single-layer) perceptron and a series of theoretical results that emphasized its limitations, where the most remarkable one is its inability to learn the XOR function \cite{minsky69perceptrons}.

This \textit{NN ice age} came to a halt in the mid 1980's, mainly with the introduction of the multi-layer perceptron (MLP) and the backpropagation algorithm \cite{Rumelhart:1986}. 
Furthermore, the revolutionary convolutional layer was presented \cite{Lecun98gradient}, where one of its notable achievements was successfully recognizing hand-written digits \cite{LeCun1990}.

While some other significant developments have happened in the following decade, such as the development of the long-short memory machine (LSTM) \cite{Hochreiter1997}, the field experienced another deceleration. 
Questions were arising with no adequate answers especially with respect to the non-convex nature of the used optimization objectives,  overfitting the training data, and the challenge of vanishing gradients. These difficulties led to a second \textit{NN winter}, which lasted two decades. 
In the meantime, classical machine learning techniques were developed and attracted much academic and industry attention. 
One of the prominent algorithms was the newly proposed Support Vector Machine (SVM) \cite{cristianini2000}, which defined a convex optimization problem with a clear mathematical interpretation~\cite{Vapnik1995SVM}. These properties increased its popularity and usage in various applications.

The $21^\text{st}$ century began with some advancements in neural networks in the areas of speech processing and Natural Language Processing (NLP). Hinton \textit{et al.}~\cite{Hinton2006} proposed a method for layer-wise initial training of neural networks, which leveraged some of the challenges in training networks with several layers.
However, the great NN \textit{tsunami} truly hit the field with the publication of \textit{AlexNet} in 2012~\cite{AlexNet}.
In this paper, Krizhevsky \textit{et al.} presented a neural network that achieved state-of-the-art performance on the ImageNet~\cite{Deng09} challenge, where the goal is to classify images into 1000 categories using 1.2 Million images for training and 150000 images for testing. The improvement over the runner-up, which relied on hand crafted features and one of the best classification techniques of that time, was notable - more than $10\%$. This caused the whole research community to understand that neural networks are way more powerful than what was thought and they bear a great potential for many applications. This led to a myriad of research works that applied NNs for various fields showing their great advantage. 

Nowadays, it is safe to say that almost every research field has been affected by this NN \textit{tsunami} wave, experiencing significant improvements in abilities and performance. 
Many of the tools used today are very similar to the ones used in the previous phase of NN. Indeed, some new regularization techniques such as batch-normalization~\cite{IoffeS15} and dropout~\cite{Srivastava2014} have been proposed. 
Yet, the key-enablers for the current success is the large amounts of data available today that are essential for large NN training, and the developments in GPU computations that accelerate the training time significantly (sometimes even leading to $\times 100$ speed-up compared to training on a conventional CPU).
The advantages of NN is remarkable especially at large scales. Thus, having large amounts of data and the appropriate hardware to process them, is vital for their success.

\begin{figure}[hb]
	\centering
	\includegraphics[width=0.75\textwidth]{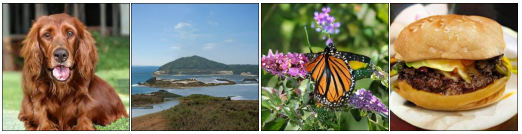}
	\caption{Class-conditional samples generated by a GAN, \cite{Brock2018}.}
	\label{fig:BigGAN_example1}
\end{figure}

A major example of a tool that did not exist before is the Generative Adversarial Network (GAN~\cite{GANs}). In 2014, Goodfellow \textit{et al.} published this novel framework for learning data distribution. The framework is composed of two models, a generator and a discriminator, trained as adversaries. The generator is trained to capture the data distribution, while the discriminator is trained to differentiate between generated (``fake'') data and real data. The goal is to let the generator synthesize data, which the discriminator fails to discriminate from the real one. The GAN architecture is used in more and more applications since its introduction in 2014. One such application is the rendering of real scene images were GANs have proved very successful~\cite{Gatys2016ImageST,Zhu2017UnpairedIT}. For example, Brock \textit{et al.} introduced the BigGAN~\cite{Brock2018} architecture that exhibited impressive results in creating high-resolution images, shown in Fig.~\ref{fig:BigGAN_example1}. While most GAN techniques learn from a set of images, recently it has been successfully demonstrated that one may even train a GAN just using one image~\cite{Shaham_2019_ICCV}.
Other GAN application include inpainting \cite{Liu_2018_ECCV,Yu_2019_ICCV}, retargeting~\cite{Shocher_2019_ICCV},
3D modeling \cite{Guibas18}, semi-supervised learning~\cite{vanEngelen2019}, domain adaptation~\cite{CyCADA2018} and more.

While neural networks are very successful, the theoretical understanding behind them is still missing. 
In this respect, there are research efforts that try to provide a mathematical formulation that explains various aspects of NN. 
For example, they study NN properties such as their optimization~\cite{sun2019optimization}, generalization \cite{Jakubovitz2019} and  expressive power \cite{Safran19Depth, ongie2020a}.

The rest of the chapter is organized as follows. In Section~\ref{sec:basic_structure} the basic structure of a NN is described, followed by details regarding popular loss functions and metric learning techniques used today (Section~\ref{sec:LF}). 
We continue with an introduction to the NN training process in Section~\ref{sec:training}, including a mathematical derivation of backpropagation and training considerations. 
Section~\ref{sec:optimizers} elaborates on the different optimizers used during training, after which Section~\ref{sec:regularizations} presents a review of common regularization schemes. Section~\ref{sec:architectures} details advanced NN architecture with state-of-the-art performances and Section~\ref{sec:summary} concludes the chapter by highlighting some current important NN challenges.

\section{Basic NN structure}\label{sec:basic_structure}
The basic building block of a NN consists of a linear operation followed by a non-linear function. 
Each building block consists of a set of parameters, termed weights and biases (sometimes the term weights includes also the biases), that are updated in the training process with the goal of minimizing a pre-defined loss function. 

\begin{figure}[hb]
	\centering
	\includegraphics[width=0.5\textwidth]{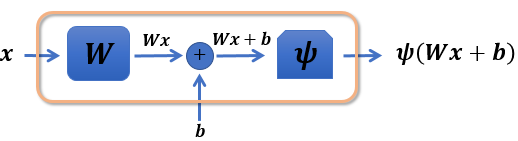}
	\caption{NN building block consists of a linear and a non-linear elements. The weights $\mathbf{W}$ and biases $\mathbf{b}$ are the parameters of the layer.}
	\label{fig:building_block}
\end{figure}

Assume an input data $\mathbf{x}\in \mathbb{R}^{d_0}$, the output of the building block is of the form \mbox{$\psi (\mathbf{W}\mathbf{x}+\mathbf{b})$}, where $\psi(\cdot )$ is a non-linear function, $\mathbf{W}\in \mathbb{R}^{d_1 \times d_0}$ is the linear operation and $\mathbf{b}\in \mathbb{R}^{d_1}$ is the bias. See Fig.~\ref{fig:building_block} for an illustration of a single building block. 
\begin{figure}[hb]
	\centering
	\includegraphics[width=\textwidth]{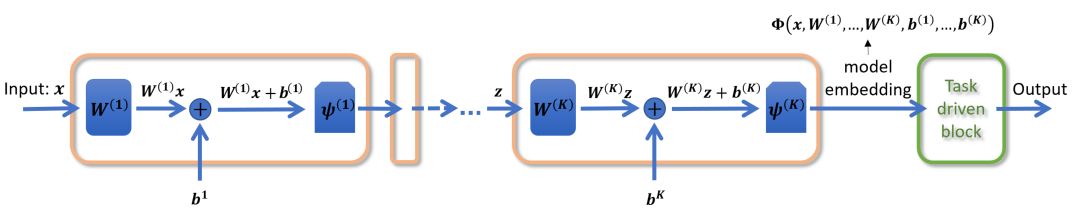}
	\caption{NN layered structure: concatenation of $N$ building blocks, e.g., model layers.}
	\label{fig:NN_illustraion}
\end{figure}

To form an NN model, such building blocks are concatenated one to another in a layered structure that allows the input data to be gradually processed as it propagates through the network. 
Such a process is termed the (feed-)forward pass. Following it, during training, a backpropagation process is used to update the NN parameters, as elaborated in Section~\ref{subsec:backprop}. In inference time, only the forward pass is used.

Fig.~\ref{fig:NN_illustraion} illustrates the concatenation of $K$ building blocks, e.g., layers. 
The intermediate output at the end of the model (before the ``task driven block'') is termed the \textit{network embedding} and it is formulated as follows:
\begin{equation}
\resizebox{.92 \textwidth}{!}{$
    \Phi(\mathbf{x},\mathbf{W}^{(1)},...,\mathbf{W}^{(K)},\mathbf{b}^{(1)},...,\mathbf{b}^{(K)})=\psi(\mathbf{W}^{(K)}...\psi(\mathbf{W}^{(2)}\psi(\mathbf{W}^{(1)}\mathbf{x}+\mathbf{b}^{(1)})+\mathbf{b}^{(2)})...+\mathbf{b}^{(K)}).
    $}
\end{equation}
The final output (prediction) of the network is estimated from the network embedding of the input data using an additional task driven layer. 
A popular example is the case of classifications, where this block is usually a linear operation followed by the \textit{cross-entropy} loss function (detailed in Section~\ref{sec:LF}). 

\begin{figure}[hb]
	\centering
	\includegraphics[width=0.35\textwidth]{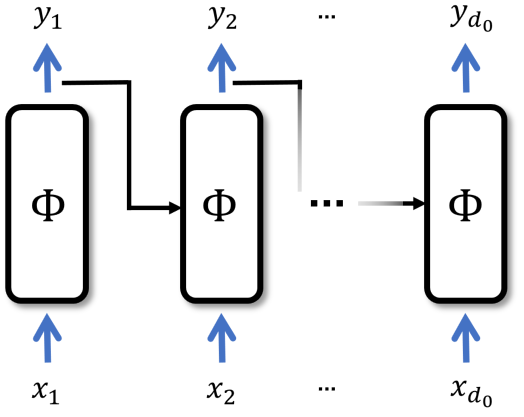}
	\caption{Recurrent NN (RNN) illustration for time series data. The feedback loop introduces time dependent characteristics to the NN model using an element-wise function. The weights are the same along all time steps.}
	\label{fig:RNN}
\end{figure}

When approaching the analysis of data with varying length, such as sequential data, a variant of the aforementioned approach is used.
A very popular example for such a neural network structure is the Recurrent Neural Network (RNN~\cite{Jain1999RNN}). 
In a vanilla RNN model, the network receives at each time step just a single input but with a feedback loop calculated using the result of the same network in the previous time-step (see an illustration in Fig.~\ref{fig:RNN}). This enables the network to "remember" information and support multiple inputs and producing one or more outputs.

More complex RNN structures include performing bi-directional calculations or adding gating to the feedback and the input received by the network. 
The most known complex RNN architecture is the Long-Term-Short-Memory (LSTM)~\cite{Hochreiter1997, gers1999learning}, which adds gates to the RNN. 
These gates decide what information from the current input and the past will be used to calculate the output and the next feedback, as well as what information to mask (i.e., causing the network to forget). This enables an easier combination of past and present information.
It is commonly used for time-series data in domains such as NLP and speech processing.

Another common network structure is the \textit{Encoder-Decoder} architecture. The first part of the model, the encoder, reduces the dimensions of the input to a compact feature vector. 
This vector functions as the input to the second part of the model, the decoder. The decoder increases its dimension, usually, back to the original input size. 
This architecture essentially learns to compress (encode) the input to an efficiently small vector and then decode the information from its compact representation. 
In the context of regular feedforward NN, this model is known as autoencoder~\cite{sonderby2016ladder} and  is used for several tasks such as image denoising \cite{Remez18Class}, image captioning \cite{Vinyals2015Show}, feature extraction~\cite{vincent2008extracting} and segmentation~\cite{atlason2019unsupervised}.
In the context of sequential data, it is used for tasks such as translation, where the decoder generates a translated sentence from a vector representing the input sentence \cite{Sutskever14Seq2Seq,cho-etal-2014-properties}.

\subsection{Common linear layers}\label{sec:layers}
\begin{figure}[hb]
	\centering
	\includegraphics[width=0.5\textwidth]{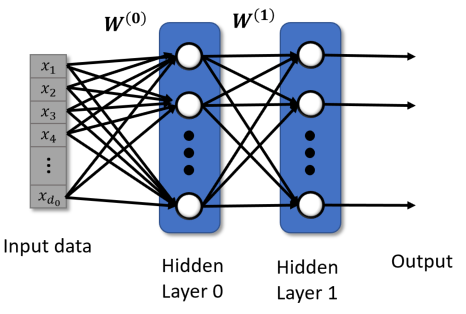}
	\caption{Fully-connected layers.}
	\label{fig:MLP}
\end{figure}

A common basic NN building block is the Fully Connected (FC) layer.
A network composed of a concatenation of such layers is termed Multi-Layer Perceptron (MLP)~\cite{Ruck1990}. 
The FC layer connects every neuron in one layer to every neuron in the following layer, i.e. the matrix $\mathbf{W}$ is dense. 
It enables information propagation from all neurons to all the ones following them. However it may not maintain spatial information. Figure~\ref{fig:MLP} illustrates a network with FC layers.

The convolutional layer~\cite{LeCun1989,Lecun98gradient} is another very common layer. We discuss here the 2D case, where the extension to other dimension is straight-forward. This layer applies one or multiple convolution filters to its input with kernels of size $W\times H$. The output of the convolution layer is commonly termed a \textit{feature map}. 

Each neuron in a feature map receives inputs from a set of neurons from the previous layer, located in a small neighborhood defined by the kernel size. If we apply this relationship recursively, we can find the part of the input that affects each neuron at a given layer, i.e., the area of visible context that each neuron sees from the input.
The size of this part is called the \textit{receptive field}. It impacts the type and size of visual features each convolution layer may extract, such as edges, corners and even patterns. 
Since convolution operations maintain spatial information and are translation equivariant, they are very useful, namely, in image processing and CV.

If the input to a convolution layer has some arbitrary third dimension, for example 3-channels in an RGB image ($C=3$) or some $C>1$ channels from an output of a hidden layer in the model, the kernel of the matching convolution layer should be of size  $W\times  H\times C$. This corresponds to applying a different convolution for each input channel separately, and then summing the outputs to create one feature map. 
The convolution layer may create a multi-channel feature map by applying multiple filters to the input, i.e., using a kernel of size \mbox{$W\times  H\times C_\text{in}\times C_\text{out}$}, where $C_\text{in}$ and $C_\text{out}$ are the number of channels at the input and output of the layer respectively.

\subsection{Common non-linear functions}\label{sec:AF}
The non-linear functions defined for each layer are of great interest since they introduce the non-linear property to the model and can limit the propagating gradient from vanishing or exploding  (see Section~\ref{sec:training}). 

\begin{figure}[hb]
	\centering
	\includegraphics[width=0.65\textwidth]{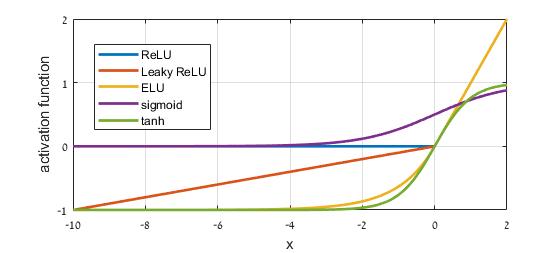}
	\caption{Different activation functions. Leaky ReLU with $\alpha=0.1$, ELU with $\alpha=1$.}
	\label{fig:activation_funcs}
\end{figure}

Non-linear functions that are applied element-wise are known as \textit{activation functions}. Common activation functions are the Rectified Linear Unit (ReLU \cite{Dahl13}), leaky ReLU \cite{Xu15}, Exponential Linear Unit (ELU) \cite{ELU15}, hyperbolic tangent (tanh) and sigmoid. 
There is no universal rule for choosing a specific activation function, however, ReLUs and ELUs are currently more popular for image processing and CV while sigmoid and tanh are more common in speech and NLP. 
Fig.~\ref{fig:activation_funcs} presents the response of the different activation functions and Table \ref{table:activation_functions} their mathematical formulation.

\begin{table}[hb]
	\centering
	\caption{Mathematical expressions for non-linear activation functions.\label{table:activation_functions}}
	\begin{tabular}{|c|c|c|c|}
	    \hline
	    \T\B Function & Formulation $s(x)$ & Derivative $\frac{ds(x)}{dx}$ & Function output range\\
		\hline
		\hline
		\T\B ReLU&  	$\begin{cases}
		0, & \text{for } x<0\\ 
		x, & \text{for } x\geq 0
		\end{cases}$
		&  $\begin{cases}
		0, & \text{for } x<0\\ 
		1, & \text{for } x\geq 0
		\end{cases}$ & $[0,\infty )$\\ \hline
		\T\B Leaky ReLU& $\begin{cases}
		\alpha x, & \text{for } x<0\\ 
		x, & \text{for } x\geq 0
		\end{cases}$ &  $\begin{cases}
		\alpha, & \text{for } x<0\\ 
		1, & \text{for } x\geq 0
		\end{cases}$& $(-\infty ,\infty )$  \\ \hline
		\T\B ELU&  $\begin{cases}
		\alpha(\mathrm{e}^{x}-1), & \text{for } x<0\\ 
		x, & \text{for } x\geq 0
		\end{cases}$&  $\begin{cases}
		\alpha \mathrm{e}^{x}, & \text{for } x<0\\ 
		1, & \text{for } x\geq 0
		\end{cases}$& $[-\alpha ,\infty )$   \\ \hline
		\T\B Sigmoid& $\frac{1}{1+\mathrm{e}^{-x}}$ &   $\frac{\mathrm{e}^{-x}}{(1+\mathrm{e}^{-x})^2}$ & $(0,1)$  \\ \hline
		\T\B tanh& $\tanh(x)=\frac{\mathrm{e}^{2x}-1}{\mathrm{e}^{2x}+1}$ &   $1-\tanh^2(x)$& $(-1,1)$   \\ \hline
	\end{tabular}
\end{table}

Another common non-linear operations in a NN model are the \textit{pooling} functions. They are aggregation operations that reduce dimensionality while keeping dominant features. 
Assume a pooling size of $q$ and an input vector to a hidden layer of size $d$, $\mathbf{z}=[z_1,z_1,...,z_d]$. For every $m\in[1,d]$, the subset of the input vector $\mathbf{\tilde{z}}=[z_m,z_{m+1},...,z_{q+m}]$ may undergo one of the following popular pooling operations:
\begin{enumerate}
    \item Max pooling: $g(\mathbf{\tilde{z}})=\max_i \mathbf{\tilde{z}}$
    \item Mean pooling: $g(\mathbf{\tilde{z}})=\frac{1}{q}\sum_{i=m}^{q+m}z_i$
    \item $\ell _p$ pooling: $g(\mathbf{\tilde{z}})=\sqrt[p]{\sum_{i=m}^{q+m} z^p_i}$
\end{enumerate}
All pooling operations are characterized by a stride, $s$, that effectively defines the output dimensions. Applying pooling with a stride $s$, is equivalent to applying the pooling with no stride (i.e., $s=1$) and then sub-sampling by a factor of $s$. It is common to add zero padding to $\mathbf{z}$ such that its length is divisible by $s$.

Another very common non-linear function is the \textit{softmax}, which normalizes vectors into probabilities. The output of the model, the embedding, may undergo an additional linear layer to transform it to a vector of size $1 \times N$, termed \textit{logits}, where $N$ is the number of classes. The logits, here denoted as $\mathbf{v}$, are the input to the softmax operation defined as follows: 
\begin{equation}\label{eq:softmax}
    \text{softmax}(v_i)=\frac{\mathrm{e}^{v_i}}{\sum_{j=1}^{N}\mathrm{e}^{v_j}}, ~~~~~ i\in[1,...,N].
\end{equation}

\section{Loss functions}\label{sec:LF}
Defining the loss function of the model, denoted as $\mathcal{L}$, is critical and usually chosen based on the characteristics of the dataset and the task at hand. 
Though datasets can vary, tasks performed by NN models can be divided into two coarse groups: (1) regression tasks and (2) classification tasks.

A {\em regression} problem aims at approximating a mapping function from input variables to a continuous output variable(s). 
For NN tasks, the output of the network should predict a continues value of interest. 
Common NN regression problems include image denoising~\cite{zhang2017beyond}, deblurring~\cite{nah2017deep}, inpainting~\cite{yang2017high} and more.
In these tasks, it is common to use the Mean Squared Error (MSE), Structural SIMilarity (SSIM) or $\ell_1$ loss as the loss function. 
The MSE ($\ell_2$ error) imposes a larger penalty for larger errors, compared to the $\ell_1$ error which is more robust to outliers in the data. 
The SSIM, and its multiscale version \cite{Zhao2017LossFF}, help improving the perceptual quality.

In the {\em classification} task, the goal is to identify the correct class of a given sample from pre-defined $N$ classes. A common loss function for such tasks is the \textit{cross-entropy} loss. 
It is implemented based on a normalized vector of probabilities corresponding to a list of potential outcomes. This normalized vector is calculated by the softmax non-linear function (Eq.~\eqref{eq:softmax}). The cross-entropy loss is defined as:
\begin{equation}\label{eq:cross-entropy}
\mathcal{L}_{CE}=-\sum_{i=1}^{N}y_i\log(p_i),
\end{equation}
where $y_i$ is the ground-truth probability (the label) of the input to belong to class $i$ and $p_i$ is the  model prediction score for this class. The label is usually binary, i.e., it contains $1$ in a single index (corresponding to the true class). This type of representation is known as \textit{one-hot encoding}. The class is predicted in the network by selecting the largest probability and the log-loss is used to increase this probability. 

Notice that a network may provide multiple outputs per input data-point. For example, in the problem of image semantic segmentation, the network predicts a class for each pixel in the image. In the task of object detection, the network outputs a list of objects, where each is defined by a bounding box (found using a regression loss) and a class (found using a classification loss). Section~\ref{subsec:detection_segmentation} details these different tasks.
Since in some problems, the labelled data are imbalanced, one may use weighted softmax (that weigh less frequent classes) or the focal loss~\cite{lin2017focal}. 

\subsection{Metric Learning}
An interesting property of the log-loss function used for classification is that it implicitly cluster classes in the network embedding space during training. However, for a clustering task, these vanilla distance criteria often produce unsatisfactory performance as different class clusters can be positioned closely in the embedding space and may cause miss-classification for samples that do not reside in the specific training set distribution.

Therefore, different metric learning techniques have been developed to produce an embedding space that brings closer intra-class samples and increases inter-class distances. This results in better accuracy and robustness of the network. It allows the network to be able to distinguish between two data samples if they are from the same class or not, just by comparing their embeddings, even if their classes have not been present at training time.

Metric learning is very useful for tasks such as face recognition and identification, where the number of subjects to be tested are not known at training time and new identities that were not present during training should also be identified/recognized (e.g., given two images the network should decide whether these correspond to the same or different persons). 

An example for a popular metric loss is the \textit{triplet loss}~\cite{facenet}. It enforces a margin between instances of the same class and other classes in the embedding feature space. This approach increases performance accuracy and robustness due to the large separation between class clusters in the embedding space.
The triplet loss can be used in various tasks, namely detection, classification, recognition and other tasks of unknown number of classes. 

In this approach, three instances are used in each training step $i$: an anchor $\mathbf{x}_i^a$, another instance $\mathbf{x}_i^p$ from the same class of the anchor (positive sample), and a sample $\mathbf{x}_i^n$ from a different class (negative class).
They are required to obey the following inequality:
\begin{equation}
    \left\Vert \Phi(\mathbf{x}_i^a)-\Phi(\mathbf{x}_i^p) \right\Vert_2^2+\alpha<\left\Vert \Phi(\mathbf{x}_i^a)-\Phi(\mathbf{x}_i^n)\right\Vert_2^2,
\end{equation}
where $\alpha<0$ enforces the wanted margin from other classes.
Thus, the triplet loss is defined by:
\begin{equation}
    \mathcal{L}=\sum_i\left\Vert \Phi(\mathbf{x}_i^a)-\Phi(\mathbf{x}_i^p)\right\Vert_2^2-\left\Vert \Phi(\mathbf{x}_i^a)-\Phi(\mathbf{x}_i^n)\right\Vert_2^2+\alpha.
\end{equation}

\begin{figure}[hb]
	\centering
	\includegraphics[width=0.6\textwidth]{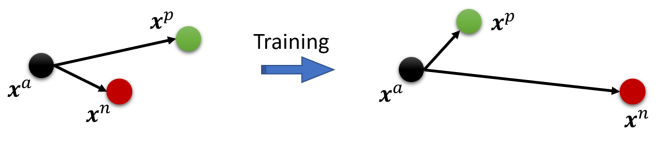}
	\caption{Triplet loss: minimizes the distance between two similar class examples (anchor and positive), and maximizes the distance between two different class examples (anchor and negative).}
	\label{fig:triplet}
\end{figure}

Fig.~\ref{fig:triplet} presents a  schematic illustration of the triplet loss influence on samples in the embedding space. 
This illustration also exhibits a specific triplet example, where the positive examples are relatively far from the anchor while negative examples are relatively near the anchor. Finding such examples that violate the triplet condition is desirable during training. They may be found by on-line or off-line searches known as \textit{hard negative mining}. A preprocessing of the instances in the embedding space is performed to find violating examples for training the network.

Finding the "best" instances for training can, evidently, aid in achieving improved convergence. However, searching for them is often time consuming and therefore alternative techniques are being explored. 

An intriguing metric learning approach relies on 'classification'-type loss functions, where the network is trained given a fixed number of classes. Yet, these losses are designed to create good embedding space that creates margin between classes, which in turn provides good prediction of similarity between two inputs. Popular examples include the Cos-loss \cite{CosFace}, Arc-loss \cite{ArcFace} and SphereFace \cite{Liu2017SphereFaceDH}.

\section{Neural network training}\label{sec:training}
Given a loss function, the weights of the neural network are updated to minimize it for a given training set. The training process of a neural network requires a large database due to the nature of the network (structure and amount of parameters) and GPUs for efficient training implementation.

In general, training methods can be divided into supervised and unsupervised training. The former consists of labeled data that are usually very expensive and time consuming to obtain. Whereas the latter is the more common case and does not assume known ground-truth labels.
However, supervised training usually achieves significantly better network performance compared to the unsupervised case. Therefore, a lot of resources are invested in labeling datasets for training. Thus, we focus here mainly on the supervised setting. 
 
In neural networks, regardless of the model task, all training phases have the same goal: to minimize a pre-defined error function, also denoted as the loss/cost function. 
This is done in two stages: (a) a feed-forward pass of the input data through all the network layers, calculating the error using the predicted outputs and their ground-truth labels (if available); followed by (b) backpropogation of the errors through the network to update their weights, from the last layer to the first. 
This process is performed continuously to find the optimized values for the weights of the network. 

The backpropagation algorithm provides the gradients of the error with respect to the network weights. These gradients are used to update the weights of the network. Calculating them based on the whole input data is computationally demanding and therefore, the common practice is to use subsets of the training set, termed \textit{mini-batches}, and cycle over the entire training set multiple times. Each cycle of training over the whole dataset is termed an \textit{epoch} and in every cycle the data samples are used in a random order to avoid biases.
The training process ends when convergence in the loss function is obtained. Since most NN problems are not convex, an optimal solution is not assured. We turn now to describe in more details the training process using backpropagation.

\subsection{Backpropogation}\label{subsec:backprop}
\begin{wrapfigure}{r}{5.5cm}
    \centering
    \includegraphics[width=0.2\textwidth]{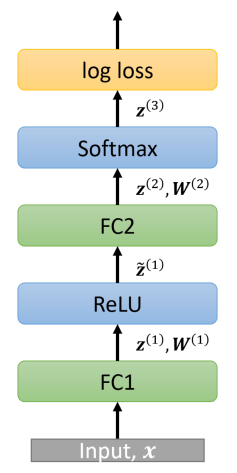}
    \caption{Simple classification model example, consisting of a two layered fully-connected model.}
    \label{fig:example}
\end{wrapfigure}

The backpropagation process is performed to update all the parameters of the model, with the goal of decreasing the loss function value. 
The process starts with a feed-forward pass of input data, $\mathbf{x}$, through all the network layers. After which the loss function value is calculated and denoted as $\Err(\mathbf{x},{\bf W})$, where ${\bf W}$ are the model parameters (including the model weights and biases, for formulation convenience). 
Then the backpropagation is initiated by computing the value of:~$\frac{\partial \Err}{\partial {\bf W}}$, followed by the update of the network weights. All the weights are updated recursively by calculating the gradients of every layer, from the final one to the input layer, using the chain rule.

Denote the output of layer $l$ as ${\bf z}^{(l)}$. Following the chain rule, the gradients of a given layer $l$ with parameters ${\bf W}^{(l)}$ with respect to its input ${\bf z}^{(l)}$ are:
\begin{equation}
    \frac{\partial \Err}{\partial {\bf z}^{(l-1)}}=\frac{\partial \Err}{\partial {\bf z}^{(l)}}\cdot\frac{\partial {\bf z}^{(l)}({\bf W}^{(l)},{\bf z}^{(l-1)})}{\partial {\bf z}^{(l-1)}},
\end{equation}
and the gradients with respect to the parameters are:
\begin{equation}
    \frac{\partial \Err}{\partial {\bf W}^{(l)}}=\frac{\partial \Err}{\partial {\bf z}^{(l)}}\cdot\frac{\partial {\bf z}^{(l)}({\bf W}^{(l)},{\bf z}^{(l-1)})}{\partial {\bf W}^{(l)}}.
\end{equation}
These two formulas of the backpropagation algorithm dictate the gradients calculation with respect to the parameters for each layer in the network and, therefore, the optimization can be performed using gradient-based optimizers (see Section~\ref{sec:optimizers} for more details).

To demonstrate the use of the backpropagation technique for the calculation of the network gradients, we turn to consider an example of a simple classification model with two-layers: a fully-connected layer with a ReLU activation function followed by another fully-connected layer with softmax function and log-loss. See Fig.~\ref{fig:example} for the model illustration.

Denote by ${\bf z}^{(3)}$ the output of the softmax layer and assume that the input $\mathbf{x}$ belongs to class $k$ (using one-hot encoding $y_k=1$). The log-loss in this case is:
\begin{equation}
    \Err=-\sum_i\log\big(z_i^{(3)}\big)y_i=-\log\Bigg(\frac{\exp\big(z^{(2)}_k\big)}{\sum_i\exp\big(z^{(2)}_i\big)}\Bigg)=-z^{(2)}_k+\log\Big(\sum_j \exp{z^{(2)}_j}\Big).
\end{equation}
For all $i\neq k$, the gradient of the error with respect to the softmax input $z_i^{(2)}$ is 
\begin{equation}
    \frac{\partial \Err}{\partial z^{(2)}_i}=\frac{\exp\big({z^{(2)}_i}\big)}{\sum_j\exp\big(z^{(2)}_j\big)}\equiv g_i.
\end{equation}
Notice that this implies that we need to decrease the value of  $z_i^{(2)}$ (the $i^{\text{th}}$-logit)  proportionally to the probability the network provides to it. 
While for the correct label, $i=k$, the derivative is:
\begin{equation}
    \frac{\partial \Err}{\partial z^{(2)}_k}=-1+\frac{\exp\big({z^{(2)}_k}\big)}{\sum_j\exp\big(z^{(2)}_j\big)}= g_k-1,
\end{equation}
which implies that the value of the logit element associated with the true label should be increased proportionally to the mistake the network is currently doing in the prediction.

The output ${\bf z}^{(2)}$ is a product of a fully-connect layer. Therefore, it can be formulated as follows:
\begin{equation}
    {\bf z}^{(2)}={\bf W}^{(2)}\tilde{{\bf z}}^{(1)},
\end{equation}
where $\tilde{{\bf z}}^{(1)}$ is the output of the ReLu function. 
Following the backpropagation rules we get that for this layer, the derivative with respect to its input is:
\begin{equation}\label{eq:backprop_fc2}
    \frac{\partial \Err}{\partial \tilde{{\bf z}}^{(1)}}=\frac{\partial \Err}{\partial {\bf z}^{(2)}}\cdot\frac{\partial {\bf z}^{(2)}({\bf W}^{(2)},\tilde{{\bf z}}^{(1)})}{\partial \tilde{{\bf z}}^{(1)}}=\frac{\partial \Err}{\partial {\bf z}^{(2)}}\cdot{\bf W}^{(2)},
\end{equation}
whereas, the derivative with respect to its parameters is:
\begin{equation}
    \frac{\partial \Err}{\partial {\bf W}^{(2)}}=\frac{\partial \Err}{\partial {\bf z}^{(2)}}\cdot\frac{\partial {\bf z}^{(2)}({\bf W}^{(2)},\tilde{{\bf z}}^{(1)})}{\partial {\bf W}^{(1)}}=\frac{\partial \Err}{\partial {\bf z}^{(2)}}\cdot \tilde{{\bf z}}^{(1)}.
\end{equation}
The ReLU operation has no weight to update, but affects the gradients. The derivative of this stage follows:
\begin{equation}
    \frac{\partial \Err}{\partial {\bf z}^{(1)}}=\frac{\partial \Err}{\partial \tilde{{\bf z}}^{(1)}}\cdot\frac{\partial \tilde{{\bf z}}^{(1)}({\bf W}^{(1)},I)}{\partial {\bf z}^{(1)}}=\begin{cases}
0, &\text{if } {\bf z}^{(1)}<0\\
\frac{\partial \Err}{\partial \tilde{{\bf z}}^{(1)}}, &\text{otherwise}.
\end{cases}
\end{equation}
The final derivative with respect to the input $\partial \Err/\partial \mathbf{x}$ is calculated similar to Eq.~\eqref{eq:backprop_fc2}.

\subsection{Training considerations}
There are several considerations that should be addressed when training a NN. 
The most infamous is the \textit{overfitting}, i.e., when the model too closely fits to the training dataset but does not generalize well to the test set. 
When this occurs, high training data precision is achieved, while the precision on the test data (not used during training) is low \cite{Tetko95}. 
For this purpose, various regularization techniques have been proposed. We discuss some of them in Section~\ref{sec:regularizations}.

A second consideration is the vanishing/exploding gradients occurring during training.
Vanishing gradients are a result of multiplications with values smaller than one during their calculation in the backpropagation recursion. This can be resolved using activation functions and batch normalization detailed in Section~\ref{sec:regularizations}.
On the other hand, the gradients might also explode due to derivatives that are significantly larger than one in the backpropogation calculation. This makes the training unstable and may imply the need for re-designing the model (e.g., replace a vanilla RNN with a gated architecture such as LSTMs) or the use of gradient clipping~\cite{Pascanu2013}.

Another important issue is the requirement that the training dataset must represent the true distribution of the task at hand. This usually enforces very large annotated datasets, which necessitate significant funding and manpower to obtain. In this case, considerable efforts must be invested to train the network using these large datasets, commonly with multiple GPUs for several days~\cite{AlexNet,karras2018progressive}. One may use techniques such as domain adaptation~\cite{Wilson2018ASO} or transfer learning~\cite{Transfer_survey} to use already existing networks or large datasets for new tasks. 

\section{Training optimizers}\label{sec:optimizers}
Training neural networks is done by applying an optimizer to reach an optimal solution for the defined loss function. 
Its goal is to find the parameters of the model, e.g., weights and biases, which achieve minimum error for the training set samples: $(\mathbf{x}_i, y_i)$, where $y_i$ is the label for the instance $\mathbf{x}_i$. For a loss function $\Err(\cdot)$, the objective reads as:
\begin{equation}
\label{eq:training_error}
    \sum_i{\Err(\Phi(\mathbf{x}_i,\mathbf{W}),y_i)},
\end{equation}
for ease of notation, all model parameters are denoted as $\mathbf{W}$.
A variety of optimizers have been proposed and implemented for minimizing Eq.~\ref{eq:training_error}. Yet, 
due to the size of the network and training dataset, mainly first-order methods are being considered, i.e. strategies that rely only on the gradients (and not on second-order derivatives such as the Hessian).

Several gradient based optimizers are commonly used for updating the parameters of the model. 
These NN parameters are updated in the opposite direction of the objective function's gradient, $g_{\{\text{GD},\mathcal{T}(t)\}}$, where $\mathcal{T}(t)$ is a randomly chosen subgroup of size $n'<n$ training samples used in iteration $t$ ($n$ is the size of the training dataset). Namely, at iteration $t$ the weights are calculated as
\begin{equation}\label{eq:update_weights}
\mathbf{W}(t)=\mathbf{W}(t-1)-\eta \cdot g_{\{\text{GD},\mathcal{T}(t)\}},
\end{equation}
where $\eta$ is the learning rate that determines the size of the steps taken to reach the (local) minimum and the gradient step, $g_{\text{\{GD},\mathcal{T}(t)\}}$ is computed using the samples in $\mathcal{T}(t)$ as
\begin{equation}\label{eq:GD}
g_{\text{\{GD},\mathcal{T}(t)\}} = \frac{1}{n'}\sum_{i\in \mathcal{T}(t)}\nabla _{W}\mathcal{L}(\mathbf{W}(t);\mathbf{x}_i;y_i),
\end{equation}
where the pair $(\mathbf{x}_i,y_i)$ is a training example and its corresponding label in the training set, and $\mathcal{L}$ is the loss function. 
However, needless to say that calculating the gradient on the whole dataset is computationally demanding. 
To this end, Stochastic Gradient Descent (SGD) is more popular, since it calculates the gradient in Eq. \eqref{eq:GD} for only one randomly chosen example from the data, i.e., $n'=1$.

Since the update by SGD depends on a different sample at each iteration, it has a high variance that causes the loss value to fluctuate. 
While this behavior may enable it to jump to a new and potentially better local minima, it might ultimately complicates convergence, as SGD may keep overshooting. 
To improve convergence and exploit parallel computing power, mini-batch SGD is proposed in which the gradient in Eq.~\eqref{eq:GD} is calculated with $n'>1$ (but not all the data).

An acceleration in convergence may be obtained by using the history of the last gradient steps, in order to stabilize the optimization. One such approach uses adaptive momentum instead of a fixed step size. This is calculated based on exponential smoothing on the gradients, i.e:
\begin{equation}
\begin{aligned}
    M(t)&=\beta\cdot M(t-1)+(1-\beta)\cdot g_{\{\text{SGD},\mathcal{T}(t)\}},\\
    \mathbf{W}(t)&=\mathbf{W}(t-1)-\eta M(t),
\end{aligned}
\end{equation}    
where $M(t)$ approximates the $1^\text{st}$ moment of $g_{\{\text{SGD},\mathcal{T}(t)\}}$. A typical value for the constant is \mbox{$\beta\sim 0.9$}, which implies taking into account the last $10$ gradient steps in the momentum variable $M(t)$~\cite{Qian99}. 
A well-known variant of Momentum proposed by Nestrov \textit{et al.}~\cite{nesterov1983method} is the Nestrov Accelerated Gradient (NAG). It is similar to Momentum but calculates the gradient step as if the network weights have been already updated with the current Momentum direction.

Another popular technique is the Adaptive Moment Estimation (ADAM)~\cite{Adam14}, which also computes adaptive learning rates. In addition to storing an exponentially decaying average of past squared gradients, $V(t)$, ADAM also keeps an exponentially decaying average of past gradients, $M(t)$, in the following way:
\begin{equation}
\begin{aligned}
M(t)&=\beta_1M(t-1)+(1-\beta_1)g_t, \\
V(t)&=\beta_2V(t-1)+(1-\beta_2)g_t^2,
\end{aligned}
\end{equation}
where $g_t$ is the gradient of the current batch, $\beta_1$ and $\beta_2$ are ADAM's hyperparameters, usually set to 0.9 and 0.999 respectively, and $M(t)$ and $V(t)$ are estimates of the first moment (the mean) and the second moment (the uncentered variance) of the gradients respectively. Hence the name of the method - Adaptive Moment Estimation.
As $M(t)$ and $V(t)$ are initialized as vectors of 0’s, the authors of ADAM observe that they are biased towards zero, especially during the initial time steps. To counteract these biases, a bias-corrected first and second moment are used: \mbox{$\hat{M}(t)=M(t)/(1-\beta_1(t))$} and \mbox{$\hat{V}(t)=V(t)/(1-\beta_2(t))$}. Therefore, the ADAM update rule is as follows:
\begin{equation}
\mathbf{W}(t+1)=\mathbf{W}(t)-\frac{\eta}{\sqrt{\hat{V}(t)+\epsilon}}\hat{M}(t).
\end{equation}
ADAM has two popular extensions: AdamW by Loshchilov \textit{et al.}~\cite{loshchilov2018decoupled} and AMSGrad by Redddi \textit{et al.}~\cite{Reddi2018}. 
There are several additional common optimizers that have adaptive momentum, such as AdaGrad \cite{Duchi2011}, AdaDelta \cite{Zeiler2012} or RMSprop \cite{Dauphin2015}.
It must be noted that since the NN optimization is non-convex, the minimal error point reached by each optimizer is rarely the same. Thus, speedy convergence is not always favored. In particular, it has been observed that Momentum leads to better generalization than ADAM, which usually converges faster~\cite{keskar2017improving}. Thus, the common practice is to make the development with ADAM and then make the final training with Momentum.

\begin{figure}[hb]
\centering
  \begin{minipage}[b]{0.495\linewidth}
    \centering
    \includegraphics[width=0.75\linewidth]{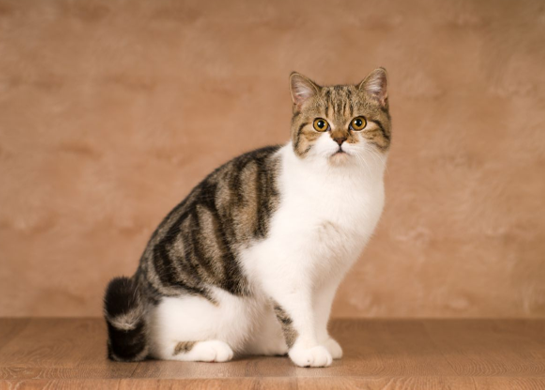} 
    \subcaption{Original image} 
    \label{fig:org} 
  \end{minipage}
  \hfill
  \begin{minipage}[b]{0.495\linewidth}
    \centering
    \includegraphics[width=0.75\linewidth]{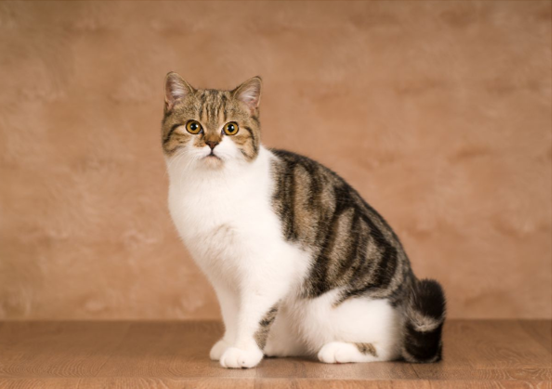} 
    \subcaption{Flip augmentation} 
    \label{fig:flip} 
  \end{minipage} 
  \\
  \begin{minipage}[b]{0.495\linewidth}
    \centering
    \begin{tabular}{cc}
    \includegraphics[width=0.45\linewidth]{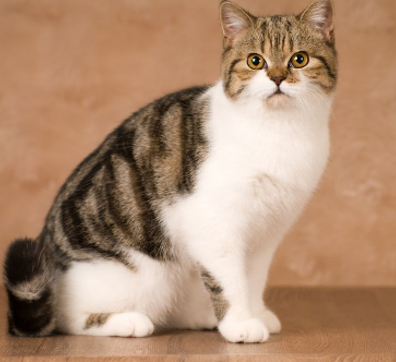} &
    \includegraphics[width=0.3\linewidth]{pics/cat_crop.png} 
    \end{tabular}
    \subcaption{Crop and scale augmentation} 
    \label{fig:crop} 
  \end{minipage}
  \hfill
  \begin{minipage}[b]{0.495\linewidth}
    \centering
    \includegraphics[width=0.75\linewidth]{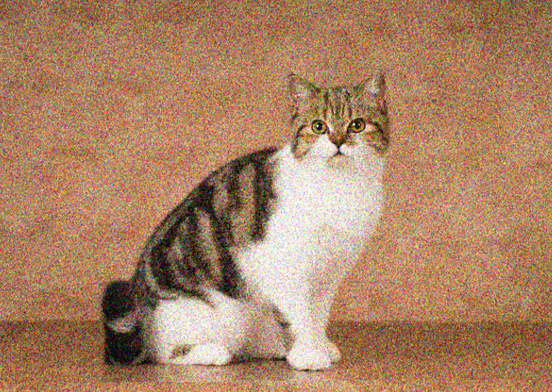} 
    \subcaption{Noise augmentation} 
    \label{fig:noise} 
  \end{minipage} 
  \caption{Different image augmentations.}
  \label{fig:augmentations} 
\end{figure}

\section{Training regularizations}\label{sec:regularizations}
One of the great advantageous of NN is their ability to generalize, i.e., correctly predict unseen data \cite{Jakubovitz2019}. This must be ensured during the training process and is accomplished by several regularization methods, detailed here. 
The most common are weight decay~\cite{Krogh_WD92}, dropout~\cite{Srivastava2014}, batch normalization~\cite{IoffeS15} and the use of data augmentation~\cite{Shorten2019}. 

\textit{Weight decay} is a basic tool to limit the growth of the weights by adding a regularization term to the cost function for large weights, which is  the sum of squares of all the weights, i.e., $\sum_i |W_i|^2$. 

The key idea in \textit{dropout} is to randomly drop units (along with their connections) from the neural network during training and thus prevent units from co-adapting too much. The percentage of dropped units is critical since a large amount will result in poor learning. Common values are $20\%-50\%$ dropped units. 

\textit{Batch normalization} is a mean to deal with changes in the distribution of the model's parameters during training. The layers need to adapt to these (often noisy) changes between instances during training. Batch normalization causes the features of each training batch to have a mean of 0 and a variance of 1 in the layer it is being applied.
To normalize a value across a batch, i.e. to batch normalize the value, the batch mean, $\mu _B$, is subtracted and the result is divided by the batch standard deviation, $\sqrt{\sigma _B^2+\epsilon}$. Note that a small constant $\epsilon$ is added to the variance in order to avoid dividing by zero. The batch normalizing transform of a given input, $\mathbf{x}$, is:
\begin{equation}
\text{BN}(\mathbf{x})=\gamma \Bigg( \frac{\mathbf{x}-\mu_B}{\sqrt{\sigma _B^2+\epsilon}}\Bigg)+\beta.
\end{equation}
Notice the (learnable) scale and bias parameters $\gamma$ and $\beta$, which provides the NN with freedom to deviate from the zero mean and unit variance.
BN is less effective when used with small batch sizes since in this case the statistics calculated per each is less accurate. Thus, techniques such as group normalization~\cite{Wu_2018_ECCV} or Filter Response Normalization (FRN)~\cite{Singh2019} have been proposed.

\textit{Data augmentation} is a very common strategy used during training to artificially ``increase'' the size of the training data and make the network robust to transformations that do not change the input label. 
For example, in the task of classification a shifted cat is still a cat; see Fig. \ref{fig:augmentations} for more similar augmentation. In the task of denoising, flipped noisy input should result in a flipped clean output. Thus, during training the network is trained also with the transformed data to improve its performance. 

Common augmentations are randomly flipping, rotating, scaling, cropping, translating, or adding noise to the data. Other more sophisticated techniques that lead to a significant improvement in network performance include mixup \cite{zhang2018mixup}, cutout~\cite{devries2017cutput} and augmentations that are learned automatically \cite{Autoaugment2018,lim2019fast,cubuk2019randaugment}.

\section{Advanced NN architectures}\label{sec:architectures}
The basic building blocks, which compose the NN model architecture, are used in frequently innovative structures.
In this section, such known architectures with state-of-the-art performance are presented, divided by tasks and data types: detection and segmentation  tasks are described in Section~\ref{subsec:detection_segmentation}, sequential data handling is elaborated in Section~\ref{subsec:sequential} and processing data on irregular grids is presented in Section~\ref{subsec:irregular_grids}. Clearly, there are many other use-cases and architectures, which are not mentioned here. 

\subsection{Deep learning for detection and segmentation}\label{subsec:detection_segmentation}
Many research works focus on detecting multiple objects in a scene, due to its numerous applications. This problem can be divided into four sub-tasks as follows, where we refer here to image datasets although the same concept can be applied to different domains as well.
\begin{enumerate}
     \item \textit{Classification and localization}: The main object in the image is detected and then localized by a surrounding bounding box and classified from a pre-known set.
    \item \textit{Object detection}: Detection of all objects in a scene that belong to a pre-known set and then classifying and providing a bounding box for each of them.
    \item \textit{Semantic segmentation}:  Partitioning the image into coherent parts by assigning each pixel in the image with its own classification label (associated with the object the pixel belongs to). For example, having a pixel-wise differentiation between animals, sky and background (generic class for all object that no class is assigned to) in an image. 
    \item \textit{Instance segmentation}: Multiple objects segmentation and classification from a pre-known set (similar to object detection but for each object all its pixels are identified instead of providing a bounding box for it).
\end{enumerate}

Today, state-of-the-art object detection performance is achieved with architectures such as Faster-RCNN~\cite{Ren2015FasterRCNN,wang2017fast}, You Only Look Once (YOLO)~\cite{Redmon2015YouOL,YOLO2,YOLO3}, Single Shot Detector (SSD)~\cite{Liu2016SSDSS} and Fully Convolutional One-Stage Object Detection (FCOS)~\cite{FCOS2019}.
The object detection models provide a list of detected bounding boxes with the class of each of them.

Segmentation tasks are mostly implemented using fully convolutional network. Known segmentation models include UNet~\cite{Unet}, Mask-RCNN~\cite{He2017MaskR} and Deeplab~\cite{deeplabv3plus2018}. These architecture have the same input/output spatial size since the output represents the segmentation map of the input image. 

Both object detection and segmentation tasks are analyzed via the Intersection over Union (IoU) metric. The IoU is defined as the ratio between the intersection area of the object's ground-truth pixels, $B_g$, with the corresponding predicted pixels, $B_p$, and the union of these group of pixels. The IoU is formulated as:
\begin{equation}
    \text{IoU}=\frac{\text{Area}\{B_g\cap B_p\}}{\text{Area}\{B_g\cup B_p\}}.
\end{equation} 
As this measure evaluate only the quality of the bounding box, a mean Average Precision (mAP) is commonly used to evaluate the models performance. 
The mAP is defined as the ratio of the correctly detected (or segmented) objects, where an object is considered to be detected correctly if there is a bounding box for it with the correct class and a  IoU greater than 0.5 (or another specified constant). 

Another common evaluation metric is the F1 score, which is the harmonic average of the precision and the recall values. See Eq.~\eqref{eq:F1_score} below. They are calculated using the following definitions that are presented for the case of semantic segmentation:
\begin{itemize}
    \item True Positive (TP): the predicted class of a pixel matches it ground-truth label.
    \item False Positive (FP): the predicted pixel of an object was falsely determined.
    \item False Negative (FN): a ground-truth pixel of an object was not predicted.
\end{itemize}
Now that they are defined, the \textit{precision}, \textit{recall} and F1 are given by:
\begin{equation}
    \text{precision}=\frac{\text{TP}}{\text{TP}+\text{FP}},\hspace{10pt}
    \text{recall}=\frac{\text{TP}}{\text{TP}+\text{FN}}
\end{equation}
\begin{equation}\label{eq:F1_score}
    \text{F1}=2\cdot\frac{\text{precision}\cdot\text{recall}}{\text{precision}+\text{recall}}.    
\end{equation}

\subsection{Deep learning on sequential data}\label{subsec:sequential}
Sequential data are composed of time-sensitive signals such as the output of different sensors, audio recordings, NLP sentences or any signal that its order is of importance. Therefore, this data must be processed accordingly. 

Initially, sequential data was processed with Recurrent NN (RNN)~\cite{Jain1999RNN} that has recurrent (feedback) connections, where outputs of the network at a given time-step serve as input to the model (in addition to the input data) at the next time-step. This introduces the time dependent feature of the NN. A RNN is illustrated in Fig.~\ref{fig:RNN}. 

However, it was quickly realized that during training, vanilla RNNs suffer from vanishing/exploding gradients. This phenomena, originated from the use of finite-precision back-propagation process, limits the size of the sequence. 

To this end, a corner stone block is used: the Long-Short-Term-Memory (LSTM~\cite{Hochreiter1997}). Mostly used for NLP tasks, the LSTM is a RNN block with gates. 
During training, these gates learn which part of the sentence to forget or to memorize. 
The gating allow some of the gradients to backpropagate unchanged, which aids the vanishing gradient symptom.
Notice that RNNs (and LSTMs) can process a sentence in a bi-directional mode, i.e., process a sentence in two directions, from the beginning to the end and vice verse. This mechanism allows a better grasp of the input context by the network.
Examples for popular research tasks in NLP data include question answering~\cite{Radford2018ImprovingLU}, translation~\cite{lample2017unsupervised} and  text generation~\cite{TextGeneration}.

{\bf Sentences processing.} An important issue in NLP is representing words in preparation to serve as network input.
The use of straight forward indices is not effective since there are thousands of words in a language. 
Therefore, it is common to process text data via \textit{word embedding}, which is a vector representation of each word in some fixed dimension. This method enables to encapsulate relationships between words. 

A classic methodology to calculate the word embedding is \textit{Word2Vec}~\cite{NIPS2013_5021}, in which these vector representations are calculated using a NN model that learn their context. 
More advanced options for creating efficient word representations include BERT~ \cite{Bert2018}, ELMO~\cite{Peters2018}, RoBERTa~\cite{liu2020roberta} and XLNet~\cite{yang2019xlnet_arxiv}.

{\bf Audio processing.} Audio recordings are used for multiple interesting tasks, such as speech to text, text to speech and speech processing. 
In the audio case, the common input to speech systems is the Mel Frequency Cepstral Coefficient (MFCC) or a Short Time Fourier Transform (STFT) image, as opposed to the audio raw-data.
A milestone example for speech processing NN architecture is the \textit{wavenet} \cite{WaveNet_Arxiv}. This architecture is an autoregressive model that synthesizes speech or audio signals. It is based on dilated convolutional layers that have large receptive fields, that allow efficient processing. Another prominent synthesis model for sequential data is the Tacotron~\cite{Shen18Natural}.

{\bf The attention model.} 
As mentioned in Section~\ref{sec:basic_structure}, one may use RNN for translation using the encoder decoder model, which encodes a source sentence into a vector, which is then decoded to a target language. 
Instead of relying on a compressed vector, which may lose information, the \textit{attention models} learn where or what to focus on from the whole input sequence. 
Introduced in 2015~\cite{Bengio2015}, attention models have shown superior performance over encoder-decoder architectures in tasks such as translation, text to speech and image captioning.
Recently, it has been suggested to replace the recurrent network structure totally by the attention mechanism, which results with the \textit{transformers network} models~\cite{Transformers2017}.

\subsection{Deep learning on irregular grids}\label{subsec:irregular_grids}
A wide variety of data acquisition mechanisms do not represent the data on a grid as is common with images data. A prominent example is 3D imaging (e.g. using LIDAR), where the input data are represented as points in a 3D space with or without color information. 
Processing such data is not trivial as standard network components, such as convolutions, assume a grid of the data. Therefore, they cannot be applied as is and custom operations are required. We focus our discussion here on the case of NN for 3D data.

Today, real-time processing of 3D scenes can be achieved with advanced NN models that are customized to these irregular grids. The different processing techniques for these irregular grid data can be divided by the type of representation used for the data:
\begin{enumerate}
    \item {\bf Points processing.} 3D data points are processed as points in space, i.e., a list of the point coordinates is given as the input to the NN. A popular network for this representation is  \textit{PointNet}~\cite{Qi2016PointNetDL}. It is the first to efficiently achieve satisfactory results directly on the point cloud. Yet, it is limited by the number of points that can be analyzed, computational time and performance. Some more recent models that improves its performance include PointNet++~\cite{qi2017pointnetplusplus}, PointCNN~\cite{Li2018PointCNNCO}, DGCNN \cite{dgcnn}. Strategies to improve its efficiency have been proposed in learning to sample~\cite{Dovrat_2019_CVPR} and RandLA-Net~\cite{hu2019randla}.
    \item {\bf Multi-view 2D projections.} 3D data points are projected (from various angles) to the 2D domain so that known 2D processing techniques can be used~\cite{NIPS2016,Kalogerakis2016}. 
    \item {\bf Volumetric (voxels).} 3D data points are represented in a grid-based \textit{voxel} representation. This is analogous to a 2D representation and is therefore advantageous. However, it is computationally exhaustive~\cite{Wu2014} and losses resolution.
    \item {\bf Meshes.} Mesh represents the 3D domain via a graph that defines the connectivity between the different points. Yet, this graph has a special structure such that it creates the surface of the 3D shape (in the common case of triangular mesh, the shape surface is presented by a set of triangles connected to each other).
    In 2015 Masci \textit{et al.}~\cite{Boscaini2015LearningCD} have shown it is possible to learn features using DL on meshes. Since then, a significant advancement has been made in mesh processing~\cite{hanocka2019meshcnn,Bronstein2017}.
    \item {\bf Graphs.} Graph representations are common for representing non-linear structured data. Some works have proposed efficient NN models for 3D data points on a grid-based graph structure~\cite{Such2017RobustSF,Niepert2016}.
\end{enumerate}

\section{Summary}\label{sec:summary}

This chapter provided a general survey of the basic concepts in neural networks. As this field is expanding very fast, the space is too short to describe all the developments in it, even though most of them are from the past eight years. 
Yet, we briefly mention here few important problems that are currently being studied. 

\begin{enumerate}
    \item {\bf Domain adaptation and transfer learning.} As many applications necessitate data that is very difficult to obtain, some methods aim at training models based on scarce datasets.
    A popular methodology for dealing with insufficient annotated data is \textit{domain adaptation}, in which a robust and high performance NN model, trained on a source distribution, is used to aid the training of a similar model (usually with the same goal, e.g., in classification the same classes are searched for) on data from a target distribution that are either  unlabelled or small in number \cite{ganin2014unsupervised,pan2010domain,DIRT-T}. 
    An example is adapting a NN trained on simulation data to real-life data with the same labels~\cite{Tzeng2017AdversarialDD, CyCADA2018}. 
    On a similar note, \textit{transfer learning}~\cite{Transfer_survey,DeCAF14} can also be used in similar cases, where in addition to the difference in the data, the input and output tasks are not the same but only similar (in domain adaptation the task is the same and only the distributions are different). One such example, is using a network trained on natural images to classify medical data~\cite{Bar15Deep}.

    \item {\bf Few shot learning.} A special case of learning with small datasets is \textit{few-shot learning}~\cite{Wang2019}, where one is provided either with just semantic information of the target classes (zero-shot learning), only one labelled example per class (1-shot learning) or just few samples (general few-shot learning). 
    Approaches developed for these problems have shown great success in many applications, such as image classification~\cite{sung2018learning, NIPS2018_7549,Sun_2019_CVPR}, object detection~\cite{Karlinsky_2019_CVPR} and segmentation~\cite{caelles2017one}.

    \item {\bf On-line learning.} Various deep learning challenges occur due to new distributions or class types introduced to the model during a continuous operation of the system (post-training), and now must be learnt by the model. The model can update its weights to incorporate these new data using \textit{on-line learning} techniques. 
    There is a need for special training in this case, as  systems that just learn based on the new examples may suffer from a reduced performance on the original data. This phenomena is known as catastrophic forgetting~\cite{kemker2018measuring}. Often, the model tends to forget the representation of part of the distribution it already learned and thus it develops a bias towards the new data. 
    A specific example of on-line learning is \textit{incremental learning}~\cite{castro2018end}, where the new data is of different classes than the original ones.   
    \item {\bf AutoML.} When approaching real-life problems, there is an inherent pipeline of tasks to be preformed before using DL tools, such as problem definition, preparing the data and processing it.
    Commonly, these tasks are preformed by specialists and require deep system understating. 
    To this end, the \textit{autoML} paradigm attempts to generalize this process by automatically learning and tuning the model used~\cite{autoML}. 
    
    A particular popular task in autoML is \textit{Neural  Architecture Search (NAS)}~\cite{Elsken2018NeuralAS}. This is of interest since the NN architecture restricts its performance. 
    However, searching for the optimal architecture for a specific task, and from a set of pre-defined operations, is computationally exhaustive when performed in a straight forward manner. 
    Therefore, on-going research attempts to overcome this limitation. An example is the DARTS~\cite{liu2018darts} strategy and its extensions~\cite{noy2019asap,chen2019progressiveDARTS} where the key contribution is finding, in a differentiable manner, the connections between network operations that form a NN architecture. This framework decreases the search time and improves the final accuracy.
    
    \item {\bf Reinforcement Learning.} To date, the most effective training method for decision based actions, such as robot movement and video games, is \textit{Reinforcement Learning} (RL)~\cite{kaelbling1996reinforcement,sutton2018reinforcement}. In RL, the model tries to maximize some pre-defined award score by learning which action to take, from a set of defined actions in specific scenarios.
\end{enumerate}

To summarize, being able to efficiently train deep neural networks has revolutionized almost every aspect of the modern day-to-day life. Examples span from bio-medical applications through computer graphics in movies and videos to international scale applications of big companies, such as Google, Amazon, Microsoft, Apple and Facebook.
Evidently, this theory is drawing much attention and we believe there is still much to unravel, including exploring and understanding the NN's potential abilities and limitations.

The next chapters detail Convolutional Neural Networks (CNN), Recurrent Neural Networks (RNN), generative models and autoencoders. All are very important paradigms that are used in numerous applications. 

\bibliographystyle{spmpsci}  
\bibliography{refs}

\end{document}